\documentclass[10pt,twocolumn,letterpaper]{article}

\usepackage{cvpr}              

\usepackage{algorithm}
\usepackage{algorithmic}
\usepackage{amsmath}        
\usepackage{bm}             
\usepackage{marvosym}       

\definecolor{gamemindaccent}{RGB}{254,60,0}
\makeatletter
\AtBeginDocument{%
  \def\@maketitle{%
    \newpage
    \null
    \vspace*{-0.2in}%
    \noindent%
    \begin{minipage}[c]{0.24\textwidth}
      \includegraphics[height=1.18cm]{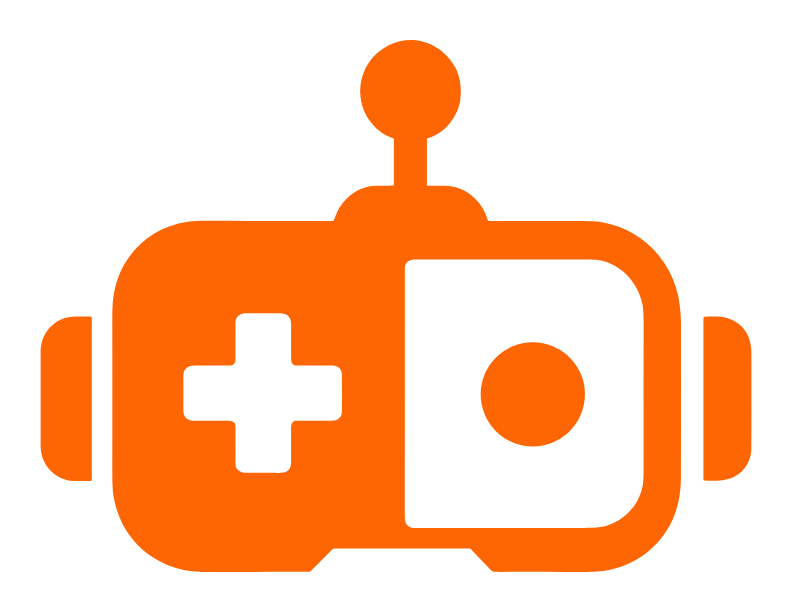}
    \end{minipage}%
    \hfill%
    \begin{minipage}[c]{0.26\textwidth}
      \raggedleft
      {\sffamily\bfseries\fontsize{15}{18}\selectfont\color{gamemindaccent} GameMind}
    \end{minipage}%
    \vspace{6pt}%
    {\color{gamemindaccent}\hrule height 1.2pt}%
    \vspace{0.55cm}%
    \iftoggle{cvprrebuttal}{\vspace*{-.3in}}{\vskip .375in}
    \begin{center}
      \iftoggle{cvprrebuttal}{{\large \bf \@title \par}}{{\Large \bf \@title \par}}
      \iftoggle{cvprrebuttal}{\vspace*{-22pt}}{\vspace*{24pt}}{
        \large
        \lineskip .5em
        \begin{tabular}[t]{c}
          \iftoggle{cvprfinal}{
            \@author
          }{
            \iftoggle{cvprrebuttal}{}{
              Anonymous \confName~submission\\
              \vspace*{1pt}\\
              Paper ID \paperID
            }
          }
        \end{tabular}
        \par
      }
      \vskip .5em
      \vspace*{12pt}
    \end{center}
  }
}
\makeatother

\definecolor{cvprblue}{rgb}{0.21,0.49,0.74}
\usepackage[pagebackref,breaklinks,colorlinks,allcolors=cvprblue]{hyperref}

\def\paperID{13086} 
\def\confName{CVPR}
\def\confYear{2026}

\title{Native3D: End-to-End 3D Scene Generation via Unified Mesh-Texture Modeling and Semantic Alignment}

\author{Yibo Liu$^{*}$, Ziwei Zhang$^{*}$, Haozhou Pang, Menghao Li, Lanshan He, Gan Qi$^{\textrm{\Letter}}$\\
Kuaishou GameMind Lab\\
{\tt\small \{liuyibo10, zhangziwei09, panghaozhou, limenghao, helanshan, qigan\}@kuaishou.com}
\thanks{Equal contribution. $^{\textrm{\Letter}}$Corresponding author.}
}

\begin{document}
\maketitle
\AddToShipoutPicture*{
    \AtPageLowerLeft{
        \put(0.5\paperwidth, \LenToUnit{1.2cm}){
            \makebox[0pt][c]{\small \textit{Accepted to CVPR 2026 Findings}}
        }
    }
}
\begin{abstract}
This paper presents \textbf{\emph{Native3D}}, the first end-to-end 3D scene generation framework that completely bypasses 2D intermediate representations. Traditional approaches typically require adapting 3D representations to the 2D domain to leverage pre-trained diffusion models, which inevitably introduces domain adaptation issues including geometric structural distortion and texture detail degradation. To address these limitations, we design a unified mesh-texture joint representation that simultaneously models both geometric structures and texture features through a Transformer-based scene encoder, effectively maintaining spatial relationships and visual consistency among objects within scenes. We further propose the 3D Representation Alignment Loss (3D REPA Loss), which employs an improved contrastive learning mechanism to align multi-level semantic representations in the latent space, significantly enhancing geometric and textural fidelity. Experimental results demonstrate that Native3D outperforms existing methods in both generation quality and editing flexibility, providing a novel solution for 3D scene editing.
\end{abstract}

\begin{figure*}
    \centering
    \includegraphics[width=1\linewidth]{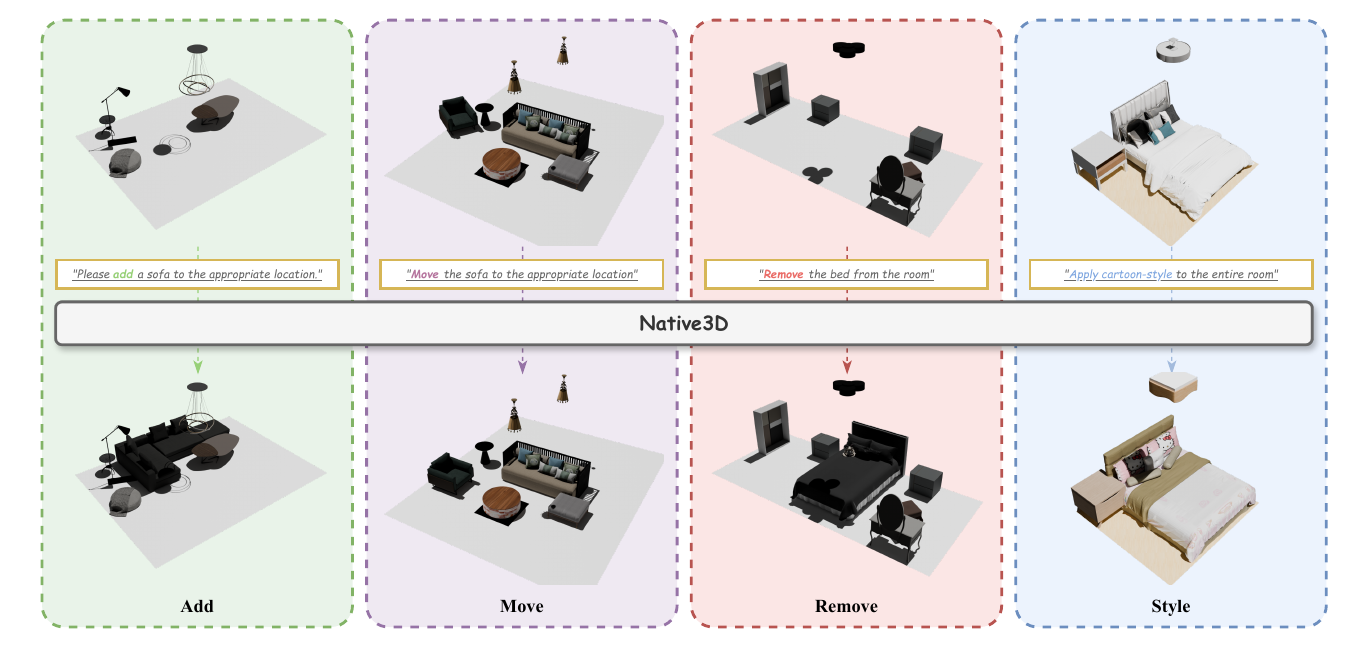}
    \caption{\emph{Native3D} supports multiple scene editing tasks: object addition, spatial rearrangement, object removal, and appearance style transfer.}
    \label{fig:head}
\end{figure*}
    
\section{Introduction}

The accelerating adoption of virtual reality (VR), augmented reality (AR), gaming, and digital media has created unprecedented demand for automated generation of high-quality 3D indoor scenes. Traditional 3D modeling workflows impose substantial barriers. These include specialized expertise, expensive software licenses, protracted production timelines, and prohibitive costs that preclude scalable deployment. Practitioners across interior design, virtual production, film industry, and architectural visualization require systems capable of synthesizing photorealistic, stylistically coherent, and geometrically detailed indoor environments from minimal input, such as textual descriptions, reference imagery, or rough sketches.

Recent advances in deep generative models present promising avenues for addressing these challenges. Large-scale pretrained text-to-image diffusion models~\cite{rombach_high-resolution_2022} exhibit remarkable proficiency in image generation and editing, establishing new paradigms for texture generation, style transfer, and content creation in architectural contexts. Concurrently, self-supervised vision transformers~\cite{caron_emerging_2021, radford_learning_2021, oquab_dinov2_2024} demonstrate superior capabilities in semantic feature extraction and cross-image correspondence, enabling more precise scene-level appearance manipulation.


Existing approaches to 3D indoor scene editing fall into two paradigms, each with fundamental limitations. The first treats scene editing as purely a 2D pixel-space problem~\cite{zhu_restyle3d_2025, huang_roompainter_2025}, leveraging semantic correspondences from pretrained diffusion models or multi-view sampling mechanisms. The core limitation of these methods lies in the absence of explicit 3D geometric constraints. 2D representations fundamentally lack the capacity to reason about 3D structures and depth relationships, resulting in cross-view inconsistencies, blurring artifacts, and visible seams under large viewpoint variations. The second paradigm adopts hybrid \emph{2D-processing-3D-reconstruction} strategies~\cite{fang_ctrl-room_2025, wang_roomtex_2024, haque_instruct-nerf2nerf_2023, tang_mvdiffusion_2023}, where inputs and outputs exist in 3D space but core operations remain anchored in 2D through iterative image editing coupled with NeRF or 3D gaussian splatting. These approaches face critical challenges in consistency propagation between 2D editing and 3D reconstruction. Lacking multi-view geometric constraints, 2D diffusion models produce cross-view inconsistencies when lifted to 3D, resulting in fine-grained misalignment and slow convergence.


These limitations converge on a central challenge: multi-view consistency, rooted in a fundamental representation mismatch. Both pure 2D methods and hybrid approaches suffer from an inherent domain gap. 2D models cannot reason about 3D structure, depth relationships, or spatial topology. Without explicit geometric constraints across viewpoints, they fail to maintain global consistency in geometry and appearance. This becomes particularly problematic when handling intricate details, preserving long-range coherence, or processing scenes with multiple objects, manifesting as cumulative errors, visual artifacts, and temporal flicker. The problem is further complicated by a persistent trade-off between efficiency and flexibility. Optimization-based techniques require known camera poses, dense captures, and substantial computation, making them impractical for sparse or casual scenarios. Forward-inference methods operating in 2D space offer speed but sacrifice 3D reasoning, struggling with occlusion handling and substantial viewpoint shifts. What the field needs are solutions that bridge this domain gap while achieving robust multi-view consistency without compromising computational practicality or editing versatility.





In this paper, we purpose \emph{Native3D}, the first end-to-end 3D-native generation framework that fundamentally circumvents the inherent domain gap issues prevalent in traditional 2D feature projection approaches. Our method directly takes 3D mesh geometry and texture information as input, achieving unified representation of global semantic information and local geometric details through a joint modeling mechanism and hierarchical feature encoder. This hybrid feature representation not only comprehensively encodes the geometric topology and visual appearance of 3D models, but also preserves precise spatial positioning information and object-level fine-grained features.

For the generation module design, we adopt Diffusion Transformer (DiT) as the core architecture, which directly learns the diffusion process in high-level 3D feature space, completely eliminating information loss caused by 2D-3D cross-domain conversion. To enable flexible scene editing capabilities, our system introduces natural language as conditional control signals, allowing users to precisely specify complex scene generation and modification requirements through textual descriptions. The output from the DiT module is reconstructed into standard 3D mesh and texture representations through a specifically designed decoder.

To further enhance the local detail fidelity of generation results, we propose the 3D REPA Loss. This loss function leverages a frozen Direct3D~\cite{wu_direct3d_2024}Encoder to extract 3D feature representations from both generated results and ground truth samples: for generated models, it extracts global features of the entire scene; for ground truth, it simultaneously extracts scene-level global features and object-level local features. By aligning these multi-scale representations in semantic feature space, the 3D REPA Loss guides the model to accurately restore geometric details and texture quality of individual target objects while maintaining global consistency.

Our work makes three primary contributions to 3D scene generation. (1) We present \emph{Native3D} that completely bypasses 2D intermediate representations, eliminating domain gap issues at the architectural level. (2) We design a unified feature representation that jointly models mesh geometry and texture appearance, enabling simultaneous capture of global layouts and local details. (3) We introduce the novel 3D REPA Loss that aligns multi-level semantic representations to significantly improve geometric and textural fidelity.

\section{Related Works}
\subsection{Text-to-3D Scene Generation}

Early text-to-3D generation methods~\cite{chen_text2shape_2018, nichol_point-e_2022, jun_shap-e_2023} primarily relied on paired text-3D datasets for training, employing GANs or latent diffusion models to learn feature representations of 3D objects. These approaches improved generation quality by scaling up training datasets, but were fundamentally limited by the scarcity of 3D data and high annotation costs, making it challenging to extend them to complex scene generation. To address the scarcity of 3D training data, recent methods~\cite{poole_dreamfusion_2022, lin_magic3d_2023, wang_prolificdreamer_2023, ouyang_text2immersion_2023, yu_viewcrafter_2024} leveraged powerful 2D text-to-image diffusion models as priors for 3D generation. Among these, DreamFusion pioneered the Score Distillation Sampling loss function, which optimized randomly initialized 3D representations through gradient descent. Magic3D further accelerated the generation process by adopting a sparse 3D hash grid structure, while ProlificDreamer effectively alleviated issues of over-saturation and low diversity through Variational Score Distillation. Despite the success of these methods in object-level generation, extending them to complete indoor scenes with layout constraints remained challenging. Methods such as Text2Room~\cite{hollein_text2room_2023} and Text2NeRF~\cite{zhang_text2nerf_2024} employed incremental generation frameworks that progressively generated different viewpoints and reconstructed 3D meshes frame by frame. However, these approaches often failed to effectively model the global layout of rooms, resulting in generated outputs that lacked spatial consistency.

\subsection{Indoor Scene Texturing and Stylization}
Several works~\cite{lin_coco-gan_2020, lin_infinitygan_2022, tang_mvdiffusion_2023, chen_scenetex_2023} represented indoor scenes as panoramic images, avoiding explicit 3D geometry modeling. These methods constructed complete panoramic views by generating and stitching multiple image patches, with MVDiffusion proposing correspondence-aware attention blocks to maintain multi-view consistency. However, such approaches could produce incorrect room layouts due to the lack of explicit geometric constraints. For the task of re-texturing given 3D scenes, RoomPainter~\cite{huang_roompainter_2025} proposed a view-integrated diffusion framework that effectively adapted 2D diffusion models to 3D-consistent texture synthesis through zero-shot techniques. This method addressed inconsistencies caused by independent view stylization and achieved high-fidelity indoor scene texture generation. While Text2Room could incrementally synthesize images and recover depth maps, it struggled with geometric and textural consistency across views, often resulting in contradictory structures.

\subsection{Semantic-Aware Appearance Transfer}

Establishing accurate semantic correspondences was crucial for achieving high-quality appearance transfer. ReStyle3D~\cite{zhu_restyle3d_2025} leveraged open-vocabulary segmentation techniques to establish dense instance-level correspondences between style and scene images, ensuring that each object was stylized with semantically matched textures. This explicit semantic correspondence mechanism could handle complex multi-object scenes and avoided semantic confusion issues present in traditional methods. Attention mechanisms in diffusion models~\cite{hertz_prompt--prompt_2022, tumanyan_plug-and-play_2022, yu_representation_2025} provided effective pathways for semantic control. Prompt-to-Prompt achieved text-based local editing by manipulating cross-attention, while Plug-and-Play maintained layout consistency using spatial features and self-attention maps. ReStyle3D further proposed a cross-image semantic attention mechanism that simultaneously transferred the appearance of all semantic categories through explicit correspondence masks, enabling efficient scene-level stylization without requiring text prompts or 3D priors.

\section{Methods}

\begin{figure*}[t]
  \centering
  \includegraphics[width=\linewidth]{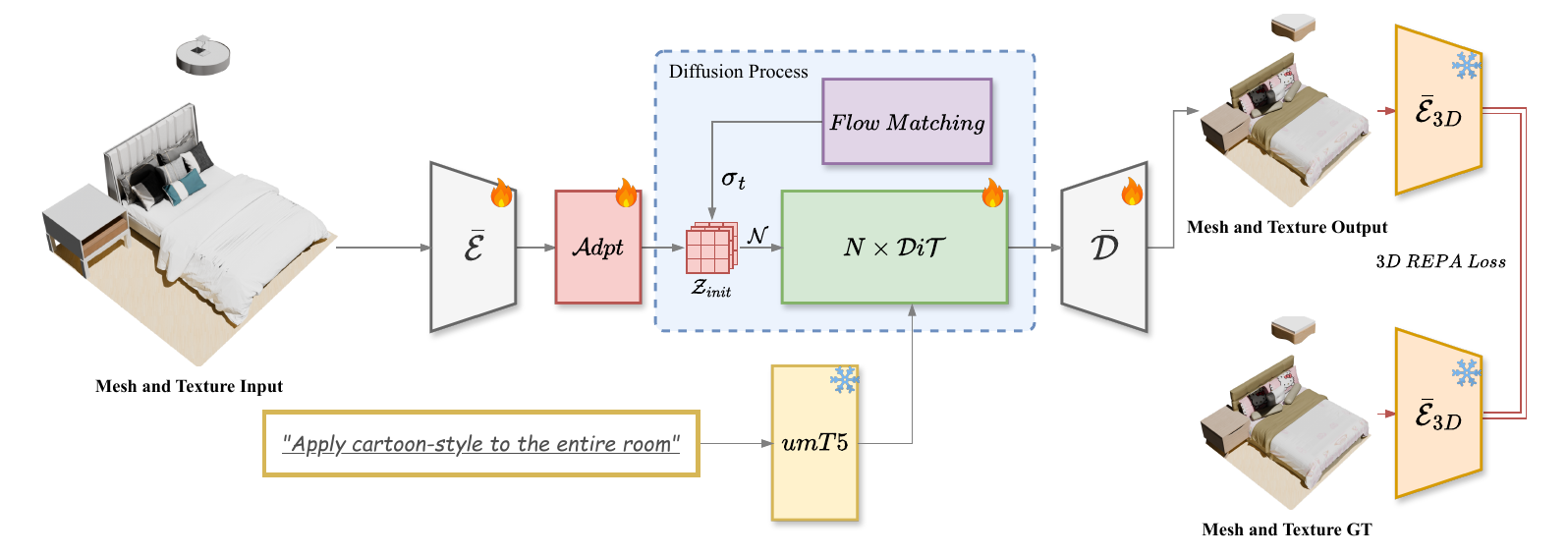}
  \caption{\textbf{Native3D Overview.} We begin by encoding the input mesh through a joint mesh-texture encoder $\bar{\mathcal{E}}$, which is initialized with the HunYuan3dShapeVAE. A trainable adapter $\mathcal{A}dpt$ then transforms the encoded latent representation to match the target dimensionality. To facilitate the diffusion process, we apply a customized flow matching scheduler that progressively corrupts the adapter output \(\mathcal{Z}_{\text{init}}\) into pure noise \(\mathcal{N}\). This noise \(\mathcal{N}\), together with text embeddings generated by the $umT5$ encoder, serves as both input and conditioning signal for the standard diffusion process. The diffusion output is subsequently decoded by the decoder $\bar{\mathcal{D}}$ to reconstruct the final mesh and texture. During training, both $\bar{\mathcal{E}}$ and $\bar{\mathcal{D}}$ are jointly optimized. We introduce an auxiliary supervision mechanism (indicated by the red line): both the model prediction and ground truth are encoded via a frozen pretrained Direct3D encoder $\bar{\mathcal{E}}_{3d}$, and their latent representations are compared using $3D\ REPA\ Loss$. This auxiliary loss mitigates detail loss inherent in joint mesh-texture modeling and contributes to the overall training objective.}

  \label{fig:Overview}
\end{figure*}

We introduce \emph{Native3D}, the first end-to-end native 3D indoor scene editing method. After reviewing foundational work in Sec.~\ref{subsec: prelim}, we present our unified scene representation framework in Sec.~\ref{subsec:unify}. Sec.~\ref{subsec:alignment} details the 3D diffusion process for scene manipulation. Finally, Sec.~\ref{subsec:loss} examines our alignment loss, which ensures geometric consistency and semantic coherence.

\subsection{Preliminaries}
\label{subsec: prelim}
\paragraph{Diffusion Models.}
Diffusion models~\cite{hertz_prompt--prompt_2022, tumanyan_plug-and-play_2022,ramesh2022hierarchical} generate data by reversing a forward process that gradually adds noise to clean samples. The forward process is typically defined as
\[
q(\mathbf{x}_t \mid \mathbf{x}_0) = \mathcal{N}(\mathbf{x}_t; \sqrt{\bar{\alpha}_t}\,\mathbf{x}_0, (1 - \bar{\alpha}_t)\mathbf{I}),
\]
where $\bar{\alpha}_t = \prod_{s=1}^t \alpha_s$ and $\alpha_t = 1 - \beta_t$, with $\beta_t$ denoting a coefficient that controls the noise strength at step $t$. Training involves learning a denoising network $\epsilon_\theta$ to predict the added noise, via the simplified objective:
\[
\mathcal{L}_{\text{simple}} = \mathbb{E}_{t, \mathbf{x}_0, \epsilon} \left[ \left\| \epsilon - \epsilon_\theta(\sqrt{\bar{\alpha}_t}\,\mathbf{x}_0 + \sqrt{1 - \bar{\alpha}_t}\,\epsilon, t) \right\|^2 \right],
\]
where $\epsilon \sim \mathcal{N}(\mathbf{0}, \mathbf{I})$. For conditional generation given text $c$, the reverse process uses $\epsilon_\theta(\cdot, t, c)$ to guide sampling.

Flow Matching (FM)~\cite{lipman2022flow} provides an alternative framework that learns a probability density path between the prior distribution $p_0(\mathbf{x})$ and the data distribution $p_1(\mathbf{x})$. It is trained to predict the velocity field $v_\theta(\mathbf{x}_t, t)$ by the ordinary differential equation: \(\frac{d\mathbf{x}_t}{dt} = v_\theta(\mathbf{x}_t, t),\)
where $\mathbf{x}_t$ is the state at time $t \in [0,1]$. The training objective minimizes the difference between the learned field and a target velocity field $u_t(\mathbf{x}_t)$:
\[
\mathcal{L}_{diff} = \mathbb{E}_{t, p_t(\mathbf{x})} \left[ \left\| v_\theta(\mathbf{x}_t, t) - u_t(\mathbf{x}_t) \right\|^2 \right].
\]
For conditional generation given a condition $c$, the vector field becomes $v_\theta(\mathbf{x}_t, t, c)$. Sampling proceeds by solving the ODE from a sample $\mathbf{x}_0 \sim p_0$:
\[
\mathbf{x}_1 = \mathbf{x}_0 + \int_0^1 v_\theta(\mathbf{x}_t, t, c)  dt,
\]
yielding a sample $\mathbf{x}_1$ distributed according to $p_{\text{data}}(\cdot \mid c)$.

\paragraph{3D Representation.}
A 3D shape is represented as a point cloud sampled from its surface, where each point combines its 3D coordinate $\mathbf{p}_i \in \mathbb{R}^3$. The input to the encoder is thus a set $\mathcal{P} = \{\mathbf{p}_i\}_{i=1}^N$.  The encoder $E$ maps this point cloud to a probabilistic latent code $\mathbf{z} \in \mathbb{R}^d$ via a variational formulation:
\[
\mathbf{z} \sim q(\mathbf{z} \mid \mathcal{P}) = \mathcal{N}(\boldsymbol{\mu}, \boldsymbol{\sigma}^2\mathbf{I}), \quad
\boldsymbol{\mu}, \boldsymbol{\sigma} = E(\mathcal{P}),
\]
where $\boldsymbol{\mu}$ and $\boldsymbol{\sigma}$ denote the predicted mean and standard deviation.
The decoder $D$ reconstructs the shape by predicting a Signed Distance Function (SDF):
\[
f_{\text{SDF}}(\mathbf{x}) = D(\mathbf{z}, \mathbf{x}), \quad \mathbf{x} \in \mathbb{R}^3,
\]
where $f_{\text{SDF}}(\mathbf{x})$ gives the signed distance from query point $\mathbf{x}$ to the nearest surface. A mesh is then extracted from the zero-level set $\{\mathbf{x} \mid f_{\text{SDF}}(\mathbf{x}) = 0\}$ using the marching cubes algorithm~\cite{lorensen1998marching}. This continuous representation enables differentiable and topology-flexible 3D reconstruction, forming the basis for latent-space generative modeling.

\subsection{Unified Scene Representation Modeling}
\label{subsec:unify}
In 3D scene editing tasks, traditional methods typically process individual objects independently, leading to results that lack global consistency and inter-object coordination. To address this challenge, we propose a unified Mesh-Texture joint representation approach that simultaneously models both the geometric structure and texture features of the entire 3D scene.

Given a 3D scene $\mathcal{S}$ containing $N$ objects, we represent it as the union of Mesh and Texture features from all objects:

\begin{equation}
\mathcal{S} = \bigcup_{i=1}^{N} (\mathcal{M}_i, \mathcal{T}_i)
\end{equation}

where $\mathcal{M}_i \in \mathbb{R}^{V_i \times 3}$ represents the mesh grid of the $i$-th object containing $V_i$ vertex coordinates, and $\mathcal{T}_i \in \mathbb{R}^{U_i \times 3}$ represents the corresponding texture features containing $U_i$ texture elements. This joint representation advantageously maintains spatial relationships and visual consistency between different objects in the scene.

\paragraph{Hierarchical Feature Encoder.}
To extract hierarchical features from 3D scenes, we design a scene encoder $E_\phi$ based on a Transformer architecture with local encoding and global aggregation modules:

\begin{equation}
\mathbf{z} = E_\phi(\mathcal{S}) = \mathrm{Adpt}\left(\mathcal{E}(\mathcal{M}_1, \mathcal{T}_1), \ldots, \mathcal{E}(\mathcal{M}_N, \mathcal{T}_N)\right)
\end{equation}

where $\mathbf{z} \in \mathbb{R}^{d_z}$ denotes the global scene feature, and $\mathcal{E}$ represents a local encoder adapted from the pre-trained HunYuan3dShapeVAE~\cite{hunyuan3d2025hunyuan3d21imageshighfidelity}. This encoder processes both mesh structures and texture information for each object. The adapter network $\mathrm{Adpt}$ employs self-attention mechanisms to model inter-object relationships and aggregate features across the scene.

\subsection{3D Representation Aligned Diffusion Process}
\label{subsec:alignment}

\paragraph{Cross-Domain Representation Adaptation.}
Many existing works utilize pre-trained 2D diffusion models by adapting 3D representations to the 2D domain. However, this cross-domain mapping inevitably introduces information loss, primarily manifested in two aspects: geometric structural distortion due to the compression of 3D spatial relationships, and texture detail degradation caused by the projection of surface properties onto 2D representations.

To address these limitations, we introduce 3D domain features into the diffusion process as a regularization alignment term. This approach preserves essential 3D structural and textural information that would otherwise be lost during domain adaptation, while still leveraging the powerful generative capabilities of pre-trained 2D diffusion models.

\paragraph{3D Feature Alignment Strategy.}
We establish feature-level alignment between the diffusion process and 3D geometric representations. Specifically, we extract rich 3D geometric features and constrain their consistency with the diffusion model's latent representations:

\begin{equation}
\min_\theta \mathcal{L}{align} = D(\mathbf{z}, \mathbf{f}{3d})
\end{equation}

where $\mathbf{z}$ denotes the diffusion latent representation, $\mathbf{f}_{3d}$ represents the extracted 3D geometric features, and $D(\cdot, \cdot)$ is a feature distance metric function. This alignment strategy ensures that the diffusion process maintains structural fidelity to the 3D domain while benefiting from 2D pre-trained models, effectively mitigating the information loss inherent in cross-domain adaptation.

\subsection{3D REPA Loss}
\label{subsec:loss}

\paragraph{Multi-Object Feature Alignment Strategy.}
Considering the unique characteristics of 3D scene editing tasks, where scene spaces typically contain multiple objects while requiring spatial continuity, we explicitly model the independent features of each object during the feature extraction stage. Assuming a scene contains $K$ objects, we obtain object-level features through a feature extractor:

\begin{equation}
\mathbf{F}_{3d} = [\mathbf{f}_{3d}^{(1)}, \mathbf{f}_{3d}^{(2)}, \ldots, \mathbf{f}_{3d}^{(K)}] \in \mathbb{R}^{K \times d_f}
\end{equation}

To facilitate alignment with the diffusion latent variables $\mathbf{z} \in \mathbb{R}^{B \times d_z}$, we merge the object dimension with the batch dimension, resulting in reshaped feature tensors $\mathbf{F}_{3d} \in \mathbb{R}^{(B \cdot K) \times d_f}$ and corresponding latent representations.

\paragraph{Improved Contrastive Alignment Loss.}
Inspired by REPA~\cite{yu_representation_2025} and DispersiveLoss~\cite{wang_diffuse_2025} from the 2D domain, we employ a modified InfoNCE loss for feature alignment. This loss imposes contrastive constraints at each level of the feature space:

\begin{equation}
\mathcal{L}_{align} = -\frac{1}{B \cdot K} \sum_{i=1}^{B \cdot K} \log \frac{\exp(s(\mathbf{z}_i, \mathbf{f}_i^+)/\tau)}{\sum_{j=1}^{B \cdot K} \exp(s(\mathbf{z}_i, \mathbf{f}_j^+)/\tau)}
\end{equation}

where $s(\cdot,\cdot)$ denotes the cosine similarity function, $\tau$ is the temperature coefficient, and $\mathbf{f}_i^+$ represents the positive sample feature corresponding to $\mathbf{z}_i$. This loss function encourages the model to maintain both the discriminative nature of individual object features and semantic consistency at the scene level.

\paragraph{Training Objective and Regularization.}
The contrastive alignment between diffusion latents and semantic features may introduce training instability due to distribution mismatches and gradient conflicts. To mitigate this issue, we incorporate latent space regularization to stabilize the training process:

\paragraph{Latent Space Regularization} imposes an $\ell_2$-norm constraint on the diffusion latent representations to ensure training stability and improve convergence:
\begin{equation}
\mathcal{L}_{reg} = \lambda |\mathbf{z}|_2^2
\end{equation}

This regularization term effectively controls the scale of latent representations and prevents excessive deviation during the alignment process, thereby maintaining numerical stability throughout training.

\paragraph{Overall Optimization Objective}
The complete training objective integrates all components through a balanced combination:
\begin{equation}
\mathcal{L}{total} = \mathcal{L}{diff} + \eta_1 \mathcal{L}{align} + \eta_2 \mathcal{L}{reg}
\end{equation}

\noindent where:
\begin{itemize}
\item $\mathcal{L}{diff}$: Standard diffusion denoising objective
\item $\mathcal{L}{align}$: Contrastive feature alignment loss
\item $\mathcal{L}_{reg}$: Latent space regularization term
\item $\eta_1, \eta_2$: Balancing hyper-parameters
\end{itemize}
\medskip
This optimized objective ensures stable training while maintaining effective feature alignment, enabling robust 3D scene editing with consistent semantic preservation.

\section{Experiments}

\begin{figure*}[t]
    \centering
    \includegraphics[width=1\linewidth]{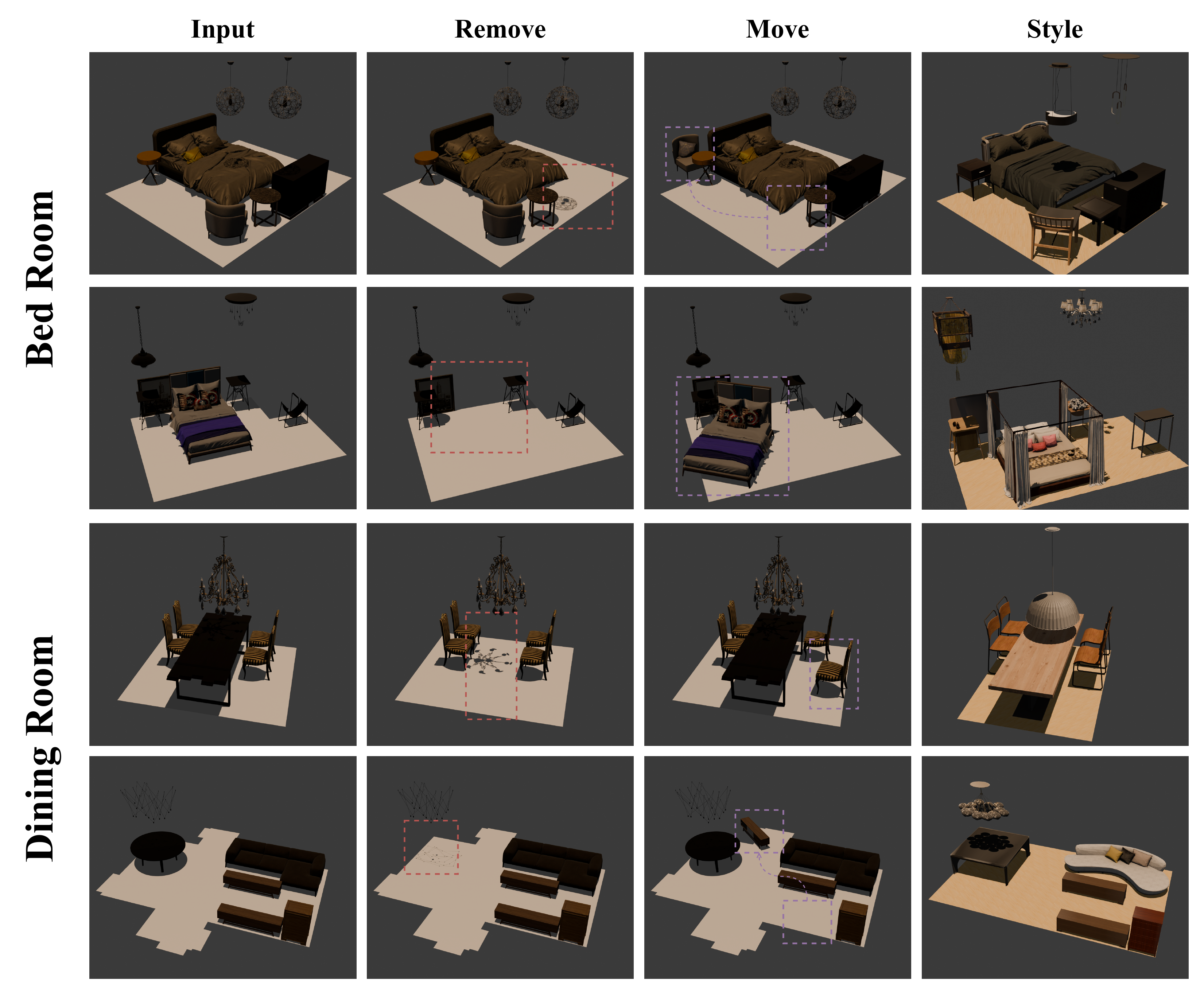}
    \caption{Qualitative editing results across two bedroom and two dining room scenes. Each row corresponds to one scene, with the first column showing the input scenes, the second demonstrating object removal, the third illustrating object movement, and the fourth showcasing style transfer.}
    \label{fig:show}
\end{figure*}

\paragraph{Dataset.}
Our collection consists of 9,001 room pairs, each accompanied by editing instructions categorized into four types: adding, removing, rearranging furniture, and style transfer. These instructions are generated using Qwen-VL-2.5~\cite{bai_qwen25-vl_2025}, ensuring precise and varied modifications. The original data is sourced from the 3D-FRONT dataset~\cite{fu_3d-front_2021}. For furniture manipulation, we utilize a custom-built Unity program, while style transfer data is synthesized utilizing InstructScene~\cite{lin_instructscene_2024}. The dataset includes various room types: 5,749 bedrooms, 653 dining rooms, 694 libraries, 305 living rooms, and 1600 other rooms. For each room type, we use 95\% of the rooms for training and 5\% for testing.

\paragraph{Evaluation Metrics.}
We employ two sets of metrics to comprehensively evaluate our 3D scene editing method. For assessing the overall quality of generated 3D scenes, we adopt commonly used quantitative metrics including CLIP Score (CS)~\cite{hessel_clipscore_2022} for measuring semantic alignment between images and text prompts, Aesthetic Score (AS)~\cite{schuhmann_laion-400m_2021} for evaluating perceptual aesthetic quality, BRISQUE (BQ)~\cite{chen2018noreferencecolorimagequality} for assessing naturalness based on locally normalized luminance, Inception Score (IS)~\cite{smith_improved_2017} for measuring quality and diversity via KL divergence, and Fréchet Inception Distance (FID)~\cite{heusel_gans_2018} for computing the distance between feature distributions of real and generated images. To evaluate fine-grained editing capabilities within 3D scenes, we leverage ImgEdit-Bench~\cite{ye_imgedit_2025}, assessing four critical tasks: \textit{Add}, \textit{Remove}, \textit{Move}, and \textit{Style}. We render edited 3D scenes from four fixed viewpoints at $0^\circ$, $30^\circ$, $60^\circ$ and $90^\circ$ azimuthal angles, and these multi-view renderings are evaluated using Qwen-VL-Max and GPT-4.1, measuring \textit{Instruction Adherence}, \textit{Editing Quality}, and \textit{Detail Preservation} on a 0-5 scale, with the final score averaged across all dimensions and viewpoints.

\subsection{Implementation Details}
In our implementation, we adopt the HunYuanDiT~\cite{zhao_hunyuan3d_2025} as diffusion generation backbone and employ Flow Matching as the sampler. For 3D shape representation, we extract geometric features using the Direct3D~\cite{wu_direct3d_2024} encoder. All experiments are conducted on high performance GPUs, with a total computational cost of approximately 3100 GPU hours. Additional implementation details are provided in the supplementary material.

\subsection{Quantitative Results}
\begin{table}[htbp]
\centering
\resizebox{\linewidth}{!}{
\begin{tabular}{cccccc}
\toprule
\textbf{Method} & \textbf{CS $\uparrow$} & \textbf{AS $\uparrow$} & \textbf{BQ $\downarrow$} & \textbf{IS $\uparrow$} & \textbf{FID $\downarrow$} \\
\midrule
\textbf{Text2tex~\cite{chen2023text2tex}} & 27.8 & 4.91 & 61.8 & 5.41 & 56.5 \\
\textbf{SceneTex~\cite{chen_scenetex_2023}} & 26.9 & 4.53 & 57.9 & 4.88 & 61.3 \\
\textbf{RoomTex~\cite{wang_roomtex_2024}} & 30.2 & 5.34 & \textbf{43.2} & 5.62 & 48.2 \\
\textbf{RoomPainter~\cite{huang_roompainter_2025}} & 31.7 & \textbf{6.12} & 47.7 & \textbf{5.81} & \textbf{35.7} \\
\textbf{Ours} & \textbf{32.1} & 5.89 & 52.8 & 5.14 & 39.8 \\
\bottomrule
\end{tabular}
}
\caption{Quantitative comparison of quality metrics (CS, AS, BQ, IS, FID). Our method achieves the best CS score and highest average ranking (1.6), demonstrating superior overall performance. While RoomPainter leads in AS, IS, FID and RoomTex excels in BQ, our approach shows exceptional consistency in semantic understanding.}
\label{tab:quality_metrics}
\end{table}

Table~\ref{tab:quality_metrics} presents the quantitative comparison using five metrics: CS, AS, BQ, IS, and FID. Our method achieves the best performance in CS and demonstrates competitive results in IS, highlighting its superior capability in semantic understanding and 3D scene representation. This advantage primarily stems from our native 3D modeling approach, which enables more accurate geometric representation and enhanced semantic comprehension. However, our method shows relatively lower performance in FID and AS, which can be attributed to its limited training on open-domain and complex datasets. Furthermore, the absence of comprehensive illumination, viewpoint, and camera parameters in our framework leads to inferior performance in BQ compared to RoomTex, which specializes in texture synthesis under such conditions. These results indicate that while our method excels in structural and semantic accuracy, there remains room for improvement in photorealistic rendering and domain generalization.

\begin{table}[htbp]
\centering
\resizebox{\linewidth}{!}{
\begin{tabular}{ccccc}
\toprule
\textbf{Method} & \textbf{Add $\uparrow$} & \textbf{Remove $\uparrow$} & \textbf{Move $\uparrow$} & \textbf{Style $\uparrow$} \\
\midrule
\textbf{Text2tex~\cite{chen2023text2tex}} & 1.45 & 2.18 & 1.11 & 3.52 \\
\textbf{SceneTex~\cite{chen_scenetex_2023}} & 2.87 & 2.99 & 1.91 & 3.66 \\
\textbf{RoomTex~\cite{wang_roomtex_2024}} & 2.66 & 2.78 & 1.82 & \textbf{4.14} \\
\textbf{RoomPainter~\cite{huang_roompainter_2025}} & 2.59 & 2.71 & 2.24 & 4.05 \\
\textbf{Ours} & \textbf{4.01} & \textbf{4.14} & \textbf{2.61} & 3.63 \\
\bottomrule
\end{tabular}
}
\caption{Quantitative comparison of editing capabilities across four fundamental operations: Add, Remove, Move, and Style. Our method achieves state-of-the-art performance on all structural editing tasks (Add, Remove, and Move), and competitively on Style editing.}
\label{tab:editing_metrics}
\end{table}

To evaluate the specific capabilities for interactive scene editing, we compare the performance on four fundamental editing operations in Table~\ref{tab:editing_metrics}: Add, Remove, Move, and Style. All metrics follow the principle that higher scores indicate better performance. The experimental results demonstrate that our proposed method achieves superior performance over all baseline models in the three core structural editing tasks: Add, Remove, and Move. This strongly validates the exceptional capability of our approach in understanding and executing complex structural modifications while maintaining scene coherence. Notably, RoomTex obtains the highest score on the Style transfer task, which can be attributed to its architecture specifically optimized for texture synthesis and style manipulation. In conclusion, these results confirm the distinct advantage of our method in handling editing tasks that require precise object manipulation and structural integrity preservation.

\begin{figure*}
    \centering
    \includegraphics[width=1\linewidth]{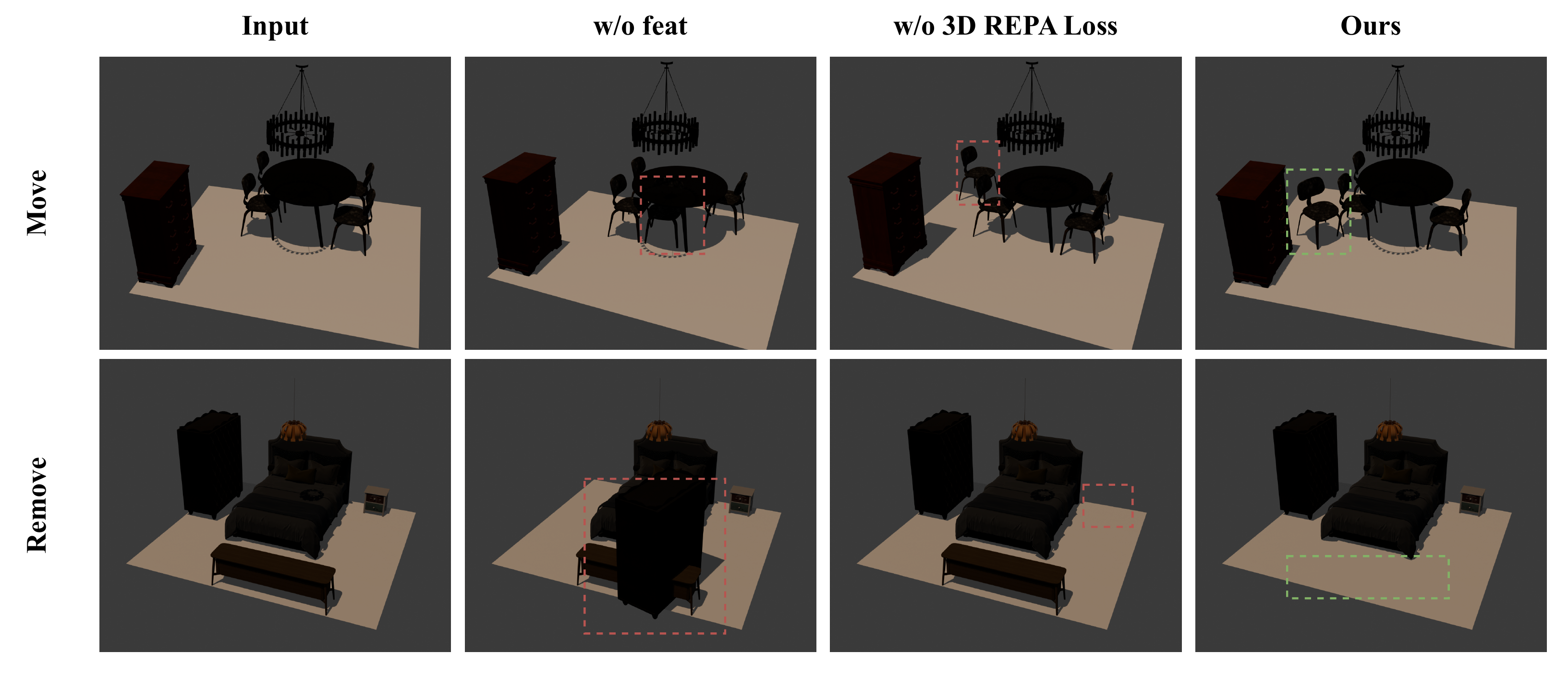}
    \caption{Ablation study on 3D feature alignment and 3D REPA loss. Move task instruction is ``Move a chair to the side of the cabinet''. Remove task instruction is ``Remove the desk''}
    \label{fig:ablation}
\end{figure*}

\subsection{Qualitative Results}
To qualitatively evaluate the editing capabilities of our method, we conducted extensive experiments across four distinct scenes - two bedroom and two dining room environments. The visual results, as shown in Fig.~\ref{fig:show}, demonstrate our approach's effectiveness in handling three fundamental editing operations: removal, movement, and style transfer. For remove operations, our method cleanly eliminates target objects while plausibly reconstructing the underlying surfaces and maintaining scene coherence. In move operations, the system successfully relocates objects to new positions with appropriate spatial relationships and lighting consistency. Most notably, the style transfer results showcase our method's ability to comprehensively alter the visual appearance of scenes - transforming both textures and materials while preserving the fundamental layout and geometry. The consistent performance across different room types and editing tasks validates the robustness and generalizability of our proposed framework.

\subsection{Ablation Study}

\begin{table}[htbp]
\centering
\resizebox{\linewidth}{!}{
\begin{tabular}{lcccc}
\toprule
\textbf{Method} & \textbf{CS $\uparrow$} & \textbf{AS $\uparrow$} & \textbf{Add $\uparrow$} & \textbf{Remove $\uparrow$} \\
\midrule
\textbf{Feature Extractor Ablation} & & & & \\
\quad 3DShape2VecSet~\cite{zhang_3dshape2vecset_2023} & 29.8 & 5.34 & 3.45 & 3.62 \\
\quad Direct3DEncoder~\cite{wu_direct3d_2024} & 31.2 & 5.67 & 3.82 & 3.95 \\
\midrule
\textbf{Loss Function Ablation} & & & & \\
\quad REPA~\cite{yu_representation_2025} & 30.5 & 5.42 & 3.61 & 3.72 \\
\quad infoNCE~\cite{oord_representation_2019} & 31.0 & 5.58 & 3.78 & 3.88 \\
\quad 3D REPA & 31.5 & 5.76 & 3.89 & 3.98 \\
\midrule
\textbf{Ours (Full)} & \textbf{32.1} & \textbf{5.89} & \textbf{4.01} & \textbf{4.14} \\
\bottomrule
\end{tabular}
}
\caption{Ablation studies on feature extractors and loss functions. We select four key metrics (CS, AS, Add, Remove) for ablation studies based on their alignment with 3D feature enhancement and semantic-spatial modeling objectives.}
\label{tab:ablation}
\end{table}

To establish a focused evaluation protocol for our ablation studies, we select CS, AS, Add, Remove as metrics. To validate the effectiveness of core components in our framework, we conduct systematic ablation studies focusing on feature extractors and loss functions. As shown in Table~\ref{tab:ablation}, Direct3DEncoder surpasses 3DShape2VecSet across all metrics, demonstrating stronger 3D understanding. Our 3D REPA loss also outperforms both REPA and infoNCE. The full model achieves the best results, validating our component design.

To more intuitively demonstrate the value of our introduced 3D feature alignment and 3D REPA loss, as shown in Fig.~\ref{fig:ablation} presents visual comparisons of the Move and Remove tasks. From the results, we can clearly observe that without incorporating 3D features to align the diffusion process, the model fails to effectively follow either type of instruction and exhibits artifacts such as penetration and visual corruption. This indicates that introducing 3D features enables more coherent modeling for generation and editing tasks. When the 3D REPA loss is absent, the edited objects either fail to comply with instructions or misinterpret the editing targets, demonstrating that additional loss supervision enhances semantic understanding and expression during the diffusion process.

\section{Conclusion}

In this paper, we present an \emph{Native3D} for indoor scene editing that fundamentally eliminates the domain gap inherent in traditional 2D projection-based approaches. By directly modeling mesh geometry and texture features through a unified representation and DiT, our method achieves comprehensive capture of both global scene layouts and fine-grained object details while enabling intuitive natural language-based control. The proposed 3D REPA Loss aligns multi-scale semantic features to preserve local geometric and textural fidelity during joint modeling. Experimental results demonstrate state-of-the-art performance in multi-view consistency and editing precision, outperforming existing 2D or hybrid methods, with promising applications in interior design, virtual reality content creation, and professional 3D design workflows.

{
    \small
    \bibliographystyle{ieeenat_fullname}
    \bibliography{CVPR26}
}


\end{document}